\def\year{2022}\relax
\documentclass[letterpaper]{article} 
\usepackage{aaai22}  
\usepackage{times}  
\usepackage{helvet}  
\usepackage{courier}  
\usepackage[hyphens]{url}  
\usepackage{graphicx} 
\usepackage{natbib}  
\usepackage{caption} 
\DeclareCaptionStyle{ruled}{labelfont=normalfont,labelsep=colon,strut=off} 
\usepackage{algorithm}
\usepackage{algorithmic}

\usepackage{newfloat}
\usepackage{listings}
\floatstyle{ruled}
\newfloat{listing}{tb}{lst}{}
\floatname{listing}{Listing}
\usepackage[sfdefault,scaled=.85]{FiraSans}
\usepackage[T1]{fontenc}
\usepackage{textcomp}
\usepackage{amsmath,amsthm}
\usepackage[cmintegrals]{newtxsf}

\usepackage[utf8]{inputenc}
\usepackage{xcolor}
\usepackage{booktabs}
\usepackage{amsthm}
\usepackage{listings}
\usepackage{inconsolata}
\usepackage{tikz}
\usepackage{pgfplots}
\usepackage{amsmath}
\usepackage{microtype}
\usepackage{algorithm}
\usepackage{cancel}

\usepackage{subcaption}
\definecolor{pixel 0}{HTML}{FFFFFF}
\definecolor{pixel 1}{HTML}{FF0000} 

\newcommand{\tw}[1]{\texttt{#1}}
\newcommand{\name}{\textsc{DCC}}
\newcommand{\popper}{\textsc{Popper}}
\newcommand{\metagol}{\textsc{Metagol}}
\newcommand{\ilasp}{\textsc{ILASP3}}
\newcommand{\ale}{\textsc{Aleph}}
\newcommand{\progol}{\textsc{Progol}}
\newcommand{\dac}{D\&C}
\newcommand{\sac}{S\&C}

\theoremstyle{definition}
\newtheorem{definition}{Definition}

\lstnewenvironment{myalgorithm}[1][] 
{
    \lstset{ 
        showspaces=false,               
        showstringspaces=false,         
        mathescape=true,
        numbers=left,
        escapeinside={*}{*},
        columns=flexible,
        numbersep=2pt,        
        keywordstyle=\bfseries,
        keywords={,and, return, not, def, in, if, else, for, foreach, while, }
        numbers=left,
        xleftmargin=.0\textwidth,
        #1 
    }
}
{}

\title{Learning Logic Programs Through Divide, Constrain, and Conquer}
\author{
    Andrew Cropper
}
\affiliations{
    University of Oxford\\
    andrew.cropper@cs.ox.ac.uk
}

\begin{document}

\maketitle

\begin{abstract}
We introduce an inductive logic programming approach that combines classical divide-and-conquer search with modern constraint-driven search.
Our anytime approach can learn optimal, recursive, and large programs and supports predicate invention.
Our experiments on three domains (classification, inductive general game playing, and program synthesis) show that our approach can increase predictive accuracies and reduce learning times.
\end{abstract}\section{Introduction}

Inductive logic programming (ILP) \cite{mugg:ilp} is a form of machine learning.
Given positive and negative examples and background knowledge (BK), the ILP problem is to find a set of rules (a \emph{hypothesis}) which  with the BK entails all the positive and none of the negative examples.

The fundamental challenge in ILP is to efficiently search a large hypothesis space (the set of all hypotheses) for a  \emph{solution} (a hypothesis that correctly generalises the examples).
Divide-and-conquer (\dac{}) approaches, such as TILDE \cite{tilde}, divide  the examples into disjoint sets and then search for a hypothesis for each set, similar to decision tree learners \cite{id3}.
Separate-and-conquer (\sac{}) approaches, such as \progol{} \cite{progol} and \ale{} \cite{aleph}, search for a hypothesis that generalises a subset of the examples, separate these examples, and then search for more rules to add to the hypothesis to generalise the remaining examples.

Although powerful, \dac{} and \sac{} approaches struggle to perform predicate invention \cite{stahl:pi} and learn recursive and optimal programs \cite{ilp30}, partly because they only consider subsets of the examples.
For instance, \progol{} and \ale{} need examples of both the base and inductive cases (in that order) to learn a recursive program.
To overcome these limitations, many modern systems learn from all examples simultaneously.
Modern systems also use powerful constraint solvers, such as answer set programming (ASP) solvers \cite{clingo}, to search for a hypothesis.
For instance, ASPAL \cite{aspal} pre-computes the set of all possible rules that may appear in a hypothesis and uses an ASP solver to find a subset that generalises the  examples.
Many modern systems can learn optimal and recursive programs but struggle to learn large programs.
For instance, \metagol{} \cite{mugg:metagold} struggles to learn programs with more than six rules and \ilasp{} \cite{ilasp3} struggles to learn rules with more than a few body literals.

To address these limitations, we introduce an approach that combines classical \dac{} search with modern constraint-driven search.
As with \dac{} approaches, we divide the examples into disjoint subsets and induce separate hypotheses for them.
Specifically, we first learn a hypothesis for each positive example.
Each hypothesis will likely be too specific.
We therefore search again for a hypothesis that generalises pairs of examples.
These new hypotheses will again likely be too specific but should generalise better than previous ones.
We repeat this process each time increasing the partition size until it matches or exceeds the number of positive examples.

The aforementioned approach on its own is pointless as it simply involves repeated search.
The key idea of our approach is to \emph{reuse} knowledge discovered when solving smaller tasks to help solve larger tasks.
For instance, when searching for a hypothesis for a single positive example, if we discover a hypothesis that incorrectly entails a negative example then any more general hypothesis will also entail it.
We can therefore ignore all generalisations of the hypothesis in subsequent iterations.

To realise our idea, we build on the constraint-driven \emph{learning from failures} (LFF) \cite{popper,poppi} ILP approach. 
The goal of LFF is to accumulate constraints to restrict the hypothesis space.
In our approach, we accumulate constraints during both the divide and conquer steps. 
In other words, we reuse constraints learned during one iteration in subsequent iterations.
We call our approach \emph{divide}, \emph{constrain}, and \emph{conquer} (\name{}).

To illustrate our approach, suppose we want to learn the following program $h_1$ to find odd elements in a list:

\begin{center}
\begin{tabular}{l}
\emph{f(A,B) $\leftarrow$ head(A,B), odd(B)}\\
\emph{f(A,B) $\leftarrow$ head(A,B), even(B), tail(A,C), f(C,B)}
\end{tabular}
\end{center}

\noindent
Suppose we have two positive examples \emph{e$_1$ = f([4,3,4,6],3)} and \emph{e$_2$ = f([2,2,9,4,8,10],9)} which correspond to odd elements in the second and third positions of a list respectively.
Also assume we have suitable negative examples and that we restrict hypotheses to definite programs \cite{lloyd:book}.
Given these examples, our approach first learns a hypothesis (in this case a single rule) for each example:
\begin{center}
\begin{tabular}{l}
\emph{$r_1$ = f(A,B) $\leftarrow$ tail(A,C), head(C,B), odd(B)}\\
\emph{$r_2$ = f(A,B) $\leftarrow$ tail(A,C), tail(C,D), head(D,B), odd(B)}
\end{tabular}
\end{center}

\noindent
These rules are too specific, i.e. they do not generalise.
Rather than stopping at this point, our approach searches again for a hypothesis $h_2$ that entails the larger chunk $\{e_1,e_2\}$.
Rather than blindly searching again, our approach reuses knowledge from the first iteration to restrict the hypothesis space.
For instance, let $|h|$ denote the number of literals in the hypothesis $h$ and assume that $r_1$ and $r_2$ are the smallest (optimal) solutions for $e_1$ and $e_2$ respectively.
Since $r_1$ and $r_2$ are optimal, the minimum size of $h_2$ is $max(|\{r_1\}|,|\{r_2\}|)$.
Likewise, since $\{r_1,r_2\} \models \{e_1,e_2\}$, we can bound the maximum size of $h_2$ as  $|\{r_1,r_2\}|$.
In this scenario of finding odd elements in a list, we can bound the size of $h_2$ as $5 \leq |h_2| < 9$, greatly reducing the hypothesis space.
We can also restrict the hypothesis space using other knowledge.
For instance, before searching for $h_2$, we can try $r_1$ and $r_2$ on $\{e_1,e_2\}$.
As neither $r_1$ nor $r_2$ alone generalises both examples, $h_2$ cannot be a specialisation of $r_1$ nor $r_2$, so we can prune all specialisations of both rules. 
As we experimentally show, this reuse of knowledge, i.e. the constrain step, is important for good learning performance. 
Given this constrained space, our approach searches again and finds the optimal solution $h_1$.

Our motivation for using a \dac{} approach is to reduce search complexity by decomposing a learning task into smaller tasks that can be solved separately.
For instance, suppose we have 10 positive examples and that 6 require a hypothesis with 7 literals; 2 require 8 literals; and 2 require 9 literals, i.e. the solution for all the examples has 24 literals.
For simplicity, consider a generate-and-test approach that enumerates all hypotheses of increasing size.
With such an approach, the search complexity $sc(h)$ of finding the hypothesis $h$ is $sc(h) = c^{|h|}$ where $c$ denotes the number of possible literals allowed in a hypothesis.
In this scenario, the complexity of finding the whole solution is $c^{24}$.
By contrast, with a \dac{} approach, we can find the individual hypotheses with the much lower cost of $c^7 + c^8 + c^9$.
Thus, our main claim is that \name{} can reduce search complexity and thus improve learning performance.

Overall, our contributions are:
\begin{itemize}
    \item We introduce a  divide, constrain, and conquer (\name{}) ILP approach.
    This anytime approach can learn optimal, recursive, and large programs and perform predicate invention.
    \item We experimentally show on three domains (classification, inductive general game playing, and program synthesis) that (i) our approach can substantially improve predictive accuracies and reduce learning times, (ii) reusing learned knowledge is vital for good learning performance, and (iii) \name{} can outperform other ILP systems.
\end{itemize}
\section{Related Work}
TILDE is a \dac{} approach.
TILDE behaves similarly to the decision tree learning algorithm C4.5 \cite{c45}.
To find a hypothesis TILDE employs a \dac{} strategy by recursively dividing the examples into disjoint subsets.
 TILDE differs from C4.5 by how it generates candidate splits to partition the examples.
C4.5 generates candidates as attribute-value pairs.
By contrast, TILDE uses conjunctions of literals (i.e. clauses/rules).
TILDE can learn large programs and can scale to large datasets.
However, it has several limitations, notably an inability to learn recursive programs, no predicate invention, and difficulty learning from small numbers of examples.

Progol is a \sac{} approach that has inspired many other approaches \cite{xhail,atom}, notably \ale{}.
Starting with an empty hypothesis, Progol picks an uncovered positive example to generalise.
To generalise an example, Progol uses mode declarations to build the \emph{bottom clause} \cite{progol}, the logically most-specific clause that explains the example.
The bottom clause bounds the search from below (the bottom clause) and above (the empty set).
Progol then uses an A* algorithm to generalise the bottom clause in a top-down (general-to-specific) manner and uses the other examples to guide the search.
Progol struggles to learn recursive and optimal programs and does not support predicate invention.
Note that Progol variants, such as \ale{} and ATOM \cite{atom}, have the same limitations.

Many recent ILP systems are \emph{meta-level} systems \cite{ilp30}.
These approaches encode the ILP problem as a meta-level logic program, i.e.~a program that reasons about programs.
Meta-level approaches often delegate the search for a solution to an off-the-shelf solver \cite{aspal,metagol,ilasp3,hexmil,dilp,apperception} after which the meta-level solution is translated back to a standard solution for the ILP task.
For instance, ASPAL translates an ILP task into a meta-level ASP program that describes every example and every possible rule in the hypothesis space.
ASPAL then delegates the search to an ASP system to find a subset of the rules that covers all the positive but none of the negative examples.
Meta-level approaches can more easily learn recursive programs and optimal programs.

A major issue with meta-level approaches is scalability.
For instance, ASPAL, HEXMIL \cite{hexmil}, and \ilasp{} \cite{ilasp3} all first pre-compute every possible rule in the hypothesis space which they pass to an ASP solver.
This approach scales well when solutions require many rules with few body literals.
However, this approach does not scale well when solutions require rules with many body literals \cite{popper}, since there are exponentially more rules given more body literals.

To improve the scalability of meta-level systems, \popper{} \cite{popper,poppi} does not precompute every possible rule in the hypothesis space.
Instead, \popper{} lazily generates rules. 
The key idea of \popper{} is to discover constraints from smaller rules (and hypotheses) to rule out larger rules.
\popper{} can learn optimal and recursive programs and perform predicate invention.
However, \popper{} searches for a single solution for all the examples and struggles to learn solutions with many literals.
As we experimentally demonstrate, our \name{} approach can substantially outperform \popper{} and other systems.

We have said that some ILP approaches struggle to learn \emph{large} programs.
However, what constitutes a large program is unclear.
Most authors measure the size of a logic program as either the number of literals \cite{ilasp3} or rules \cite{mugg:metagold} in it.
However, these two metrics are too simple.
For instance, many approaches can easily learn programs with lots of clauses by simply memorising the examples.
Likewise, approaches based on \emph{inverse entailment} \cite{progol} can easily learn programs with lots of literals by simply returning the \emph{bottom clause}.
In this paper, we do not formally define what constitutes a large program.
By \emph{large}, we informally mean programs with reasonably large numbers of rules, variables, and literals. 
\section{Problem Setting}
\label{sec:setting}

Our problem setting is the \emph{learning from failures} (LFF) \cite{popper} setting.
LFF uses \emph{hypothesis constraints} to restrict the hypothesis space.
Let $\mathcal{L}$ be a language that defines hypotheses, i.e.~a meta-language.
For instance, consider a meta-language formed of two literals \emph{h\_lit/4} and \emph{b\_lit/4} which represent \emph{head} and \emph{body} literals respectively.
With this language, we can denote the clause \emph{last(A,B) $\leftarrow$ tail(A,C), head(C,B)} as the set of literals \emph{\{h\_lit(0,last,2,(0,1)), b\_lit(0,tail,2,(0,2)), b\_lit(0,head,2,(2,1))\}}.
The first argument of each literal is the clause index, the second is the predicate symbol, the third is the arity, and the fourth is the literal variables, where \emph{0} represents \emph{A}, \emph{1} represents \emph{B}, etc.
A \emph{hypothesis constraint} is a constraint (a headless rule) expressed in $\mathcal{L}$.
Let $C$ be a set of hypothesis constraints written in a language $\mathcal{L}$.
A set of definite clauses $H$ is \emph{consistent} with $C$ if, when written in $\mathcal{L}$, $H$ does not violate any constraint in $C$.
For instance, the constraint
\emph{$\leftarrow$ h\_lit(0,last,2,(0,1)), b\_lit(0,last,2,(1,0))}
would be violated by the definite clause \emph{last(A,B) $\leftarrow$ last(B,A)}.
We denote as $\mathcal{H}_{C}$ the subset of the hypothesis space $\mathcal{H}$ which does not violate any constraint in $C$.

We define the LFF problem:

\begin{definition}[\textbf{LFF input}]
\label{def:probin}
The \emph{LFF} input is a tuple $(E^+, E^-, B, \mathcal{H}, C)$ where $E^+$ and $E^-$ are sets of ground atoms denoting positive and negative examples respectively; $B$ is a definite program denoting background knowledge;
$\mathcal{H}$ is a hypothesis space, and $C$ is a set of hypothesis constraints.
\end{definition}

\noindent
We define a LFF solution:

\begin{definition}[\textbf{LFF solution}]
\label{def:solution}
Given an input tuple $(E^+, E^-, B, \mathcal{H}, C)$, a hypothesis $H \in \mathcal{H}_{C}$ is a \emph{solution} when $H$ is \emph{complete} ($\forall e \in E^+, \; B \cup H \models e$) and \emph{consistent} ($\forall e \in E^-, \; B \cup H \not\models e$).
\end{definition}

\noindent
If a hypothesis is not a solution then it is a \emph{failure}.
A hypothesis is \emph{incomplete} when $\exists e \in E^+, \; H \cup B \not \models e$.
A hypothesis is \emph{inconsistent} when $\exists e \in E^-, \; H \cup B \models e$.
A hypothesis is \emph{totally incomplete} when $\forall e \in E^+, \; H \cup B \not \models e$.

Let $cost : \mathcal{H} \mapsto R$ be an arbitrary cost function that measures the cost of a hypothesis.
We define an \emph{optimal} solution:

\begin{definition}[\textbf{Optimal solution}]
\label{def:opthyp}
Given an input tuple $(E^+, E^-, B, \mathcal{H}, C)$, a hypothesis $H \in \mathcal{H}_{C}$ is \emph{optimal} when (i) $H$ is a solution, and (ii) $\forall H' \in \mathcal{H}_{C}$, where $H'$ is a solution, $cost(H) \leq cost(H')$.
\end{definition}

\noindent
In this paper, our cost function is the number of literals in the hypothesis $H$.


\paragraph{Constraints.}
The goal of an LFF learner is to learn hypothesis constraints from failed hypotheses.
\citet{popper,poppi} introduce hypothesis constraints based on subsumption \cite{plotkin:thesis}.
A clause $C_1$ \emph{subsumes} a clause $C_2$ ($C_1 \preceq C_2$) if and only if there exists a substitution $\theta$ such that $C_1\theta \subseteq C_2$.
A clausal theory $T_1$ subsumes a clausal theory $T_2$ ($T_1 \preceq T_2$) if and only if $\forall C_2 \in T_2, \exists C_1 \in T_1$ such that $C_1$ subsumes $C_2$.
A clausal theory $T_1$ is a \emph{specialisation} of a clausal theory $T_2$ if and only if $T_2 \preceq T_1$.
A clausal theory $T_1$ is a \emph{generalisation} of a clausal theory $T_2$ if and only if $T_1 \preceq T_2$.
If a hypothesis $H$ is incomplete, a \emph{specialisation} constraint prunes specialisations of $H$, as they are guaranteed to also be incomplete.
If a hypothesis $H$ is inconsistent, a \emph{generalisation} constraint prunes generalisations of $H$, as they are guaranteed to be inconsistent as well.
If a hypothesis $H$ is totally incomplete, a \emph{redundancy} constraint prunes hypotheses that contain a specialisation of $H$ as a subset.

\section{\name{} Algorithm}
\label{sec:impl}

We now describe our \name{} algorithm.
We first briefly describe \popper{}, which we use as our underlying search algorithm.

\subsection{\popper{}}

\noindent
Algorithm \ref{alg:popper} shows the \popper{} algorithm, which solves the LFF problem (Definition \ref{def:probin}).
\popper{} takes as input background knowledge (\tw{bk}), positive (\tw{pos}) and negative (\tw{neg}) examples, a set of hypothesis constraints (\tw{in\_cons}), and lower (\tw{min\_m}) and upper (\tw{max\_m}) bounds on hypothesis sizes.
\popper{} uses  a \emph{generate}, \emph{test}, and \emph{constrain} loop to find a solution.

\popper{} starts with a ASP program $\mathcal{P}$ (hidden in the generate function) whose models correspond to hypotheses (definite programs).
\popper{} augments $\mathcal{P}$ with ASP constraints to eliminate models and thus prune hypotheses.
In the generate stage (line 5), \popper{} uses Clingo \cite{clingo}, an ASP system, to search for a model (a hypothesis) of $\mathcal{P}$.
A constraint ensures that the hypothesis has exactly \tw{m} literals.
If a model is found, \popper{} converts it to a hypothesis (\tw{h}).
Otherwise; \popper{} increments the hypothesis size (line 7) and loops again.

If there is a hypothesis then in the test stage (line 9), \popper{} tests it on the given training examples.
If a hypothesis fails, i.e. is \emph{incomplete} or \emph{inconsistent}, then in the constrain stage (line 12), \popper{} deduces hypothesis constraints (represented as ASP constraints) from the failure which it adds to the set of constraints, which are in turn added to $\mathcal{P}$ to prune models and thus restrict the hypothesis space.
For instance, if a hypothesis is incomplete, i.e. does not entail all the positive examples, then \popper{} builds a specialisation constraint to prune hypotheses that are logically more specific.

To find an optimal solution (i.e. one with the minimum number of literals), \popper{} progressively increases the number of literals allowed in a hypothesis when the hypothesis space is empty at a certain size (e.g. when $\mathcal{P}$ has no more models).
This loop repeats until either (i) \popper{} finds an optimal solution, or (ii) there are no more hypotheses to test.

\begin{algorithm}[t]
{
\begin{myalgorithm}[]
def $\text{popper}$(bk, pos, neg, in_cons, min_m, max_m):
  cons = in_cons
  m = min_m
  while m $\leq$ max_m:
    h = generate(cons, m)
    if h == UNSAT:
      m += 1
    else:
      outcome = test(pos, neg, bk, h)
      if outcome == (COMPLETE, CONSISTENT)
        return h, cons
      cons += constrain(h, outcome)
  return {}, cons
\end{myalgorithm}
\caption{
\popper{}
}
\label{alg:popper}
}
\end{algorithm}

\subsection{\name{}}

Algorithm \ref{alg:dcc} shows the \name{} algorithm.
Before describing it in detail, we describe it at a high-level.
The idea is to divide the positive examples into \emph{chunks} of size $k$.
In the first iteration $k=1$, each example is in its own chunk.
\name{} enumerates the chunks and calls \popper{} to find a hypothesis for the chunk examples and \emph{all} the negative examples.
After enumerating all the chunks, \name{} forms an \emph{iteration} hypothesis as the union of all the chunk hypotheses.
\name{} then doubles the chunk size and repeats the process until the chunk size exceeds the number of examples.
Without optimisations, \name{} performs $\ln n$ iterations where $n$ is the number of positive examples.

We now describe \name{} in detail.
\name{} takes as input background knowledge (\tw{bk}), and positive (\tw{pos}) and negative (\tw{neg}) examples.
It maintains a set of constraints (\tw{cons}) that is initially empty (line 2).
\name{} divides the positive examples into \emph{chunks}.
In the first iteration, each positive example is in its own chunk (line 5).
The while loop in Algorithm \ref{alg:dcc} builds a hypothesis for a given chunk size.
Line 8 creates an empty hypothesis (\tw{iteration\_hs}) for the current chunk size.
\name{} divides the chunks into smaller chunks (\tw{chunk}) of size \tw{k} and enumerates them.

In line 10 \name{} calls the function \tw{lazy\_check}.
We delay discussion of this function until we have described the main \name{} loop but at a high-level this function tries to find an already discovered hypothesis that covers the current chunk.
The purpose is to reduce the number of calls to \popper{}.

Ignoring the lazy check, \name{} tries to find a hypothesis for this chunk.
The first step (line 12) selects only the relevant constraints for the examples in the chunk (\tw{chunk\_cons}).
All generalisation constraints are selected.
Specialisation constraints are only selected if they hold for at least one example in the chunk.
Redundancy constraints are selected only if they hold for all examples in the chunk.
The second step (line 13) deduces hypothesis size bounds for the subsequent \popper{} search.
The minimum size is the largest best solution for each chunk example.
The maximum size is the size of the union.

\name{} calls \popper{} with the filtered constraints and hypothesis size bounds. 
\popper{} returns a hypothesis (\tw{h}) that entails the chunk examples (if one exists) and a set of new constraints (\tw{new\_cons}).
\name{} updates its constraints (line 15) with the new ones.
If a hypothesis is found, \name{} adds it to the iteration hypothesis (line 17) and updates a hash table that maintains the last hypothesis for each example (line 18).

After passing through all the chunks, \name{} forms a single iteration hypothesis as the union of all the hypotheses (line 19).
\name{} calculates the score of this iteration hypothesis on \emph{all} the examples.
We calculate the score as the number of correctly generalised examples (true positives + true negatives).
In future work, we will explore alternative scoring functions, such as those that minimise description length \cite{mdl}.
If the score improves on the best score, \name{} updates the best hypothesis (line 22).
Line 22 \emph{compresses} the chunks.
We delay discussion of this function until we have described the main \name{} loop but at a high-level this function tries to merge examples to reduce the number of iterations.
After enumerating all the chunks, \name{} doubles the chunk size (line 24) and repeats the process.
Once the loop has finished, \name{} returns the best hypothesis.

\begin{algorithm}[t]
{
\small
\begin{myalgorithm}[]
def $\text{dcc}$(bk, pos, neg):
  cons = {}
  k = 1
  best_h, score = None, 0
  all_chunks = {{x} | x in pos}
  exs_h = {}
  while k $\leq$ |all_chunks|:
    iteration_hs = {}
    for chunk in divide(all_chunks, k):
      h = lazy_check(bk, chunk, iteration_hs)
      if h == None:
          chunk_cons = filter_c(cons, chunk)
          min_m, max_m = calc_bounds(exs_h, chunk)
          h, new_cons  = popper(bk, chunk, neg, chunk_cons, min_m, max_m)
          cons += new_cons
      if h != None:
        iteration_hs += h
        exs_hs = update(exs_hs, h, chunk)
    iteration_h = union(iteration_hs)
    h_score = test(pos, neg, bk, iteration_h)
    if h_score > best_score:
        best_h, best_score = iteration_h, h_score
    all_chunks = compress(iteration_hs, all_chunks)
    k += k
  return best_h
\end{myalgorithm}
\caption{
\name{}
}
\label{alg:dcc}
}
\end{algorithm}


\paragraph{Laziness.}
The goal of \emph{laziness} is to reduce the number of calls to \popper{}.
Suppose we have $n$ chunks of examples $e_1, e_2,\dots, e_n$ and that during the for loop in Algorithm \ref{alg:dcc} we find the solution $h_1$ for $e_1$.
We now want to find a solution $h_2$ for $e_2$.
Suppose that $h_1$ is a solution for $e_2$.
Then do we need to search for $h_2$?
On the one hand, $h_1$ may be sub-optimal in that there may be a smaller hypothesis that entails $e_2$.
On the other hand, since we need $h_1$ (or a generalisation of it) to entail $e_1$, we may as well reuse $h_1$ as it requires adding no more literals to our iteration hypothesis.
Laziness generalises to all previously found hypotheses.
Without laziness, for a chunk with $n$ examples, \name{} requires in the best- and worst-cases $n$ calls to \popper{}.
Laziness reduces the best-case to 1.
Moreover, we can deduce additional constraints from these lazy checks.
Suppose that $h_1$ does not entail $e_2$. 
We can therefore rule out all specialisations of $h_1$ when searching for $h_2$ to further restrict the hypothesis space.
In Section \ref{sec:exp}, we experimentally evaluate the impact of laziness on learning performance.

\paragraph{Compression.}
Given $n$ positive examples $e_1, e_2, \dots, e_n$, Algorithm \ref{alg:dcc} searches for solutions $h_1, h_2, \dots, h_n$ respectively.
It then increases the chunk size and searches for solutions $h_1', h_2',\dots, h{_{n/2}}'$ for $\{e_1,e_2\}, \{e_3,e_4\},\dots, \{e_{n-1},e_{n}\}$ and so on.
This approach requires in the best- and -worst-cases $\log n$ iterations and, without laziness, $n \log n$ calls to \popper{}.
Suppose that after the first iteration we know that $h_1$ entails $\{e_1\}$, $\{e_6\}$, and $\{e_9\}$; $h_2$ entails $\{e_3\}$ and $\{e_4\}$; and $h_3$ entails $\{e_2\}$, $\{e_5\}$, and $\{e_{13}\}$.
Then in the second iteration Algorithm \ref{alg:dcc} will search for solutions $h_1', h_2', \dots, h_{n/2'}$ etc.
The two searches for $h_1'$ to cover $\{e_1,e_2\}$ and $h_3'$ to cover $\{e_5, e_6\}$ are basically the same.
To reduce the number of iterations, we can \emph{compress} the chunks by the hypotheses that entail them.
In other words, we can merge two chunks if they are covered by the same hypothesis.
In the above case, we create a $h_1$ bucket with $\{e_1,e_6,e_9\}$, a $h_2$ bucket with $\{e_3,e_4\}$, and a $h_3$ bucket with $\{e_2,e_5,e_{13}\}$. 
With this compression approach in the second iteration, we now search for $h_1'$ for the chunk $\{e_1,e_6,e_9,e_3,e_4\}$ (buckets 1 and 2) and $h_2'$ for the chunk $\{e_2,e_5,e_{13}\}$.
Compression reduces the best-case number of iterations from $\log n$ to $1$ and the best-case number of calls (without laziness) to \popper{} from $n \log n$ to $n$. 
In Section \ref{sec:exp}, we experimentally evaluate the impact of compression on learning performance.

\paragraph{Anytime.}
\name{} is an anytime algorithm.
If at any point a user stops the search or the search duration exceeds a timeout, \name{} returns the best hypothesis thus far.\section{Experiments}
\label{sec:exp}

We claim that \name{} can reduce search complexity and thus improve learning performance.
To evaluate this claim, our experiments aim to answer the question:

\begin{description}
\item[Q1] Can \name{} improve predictive accuracies and reduce learning times?
\end{description}

\noindent
To answer \textbf{Q1}, we compare \name{} against \popper{}.
This comparison allows us to answer \textbf{Q1} as \name{} uses \popper{} as its underlying search algorithm and both systems use identical biases.
Comparing against other systems will not allow us to answer \textbf{Q1}.

\name{} has various optimisations that we claim improve learning performance, notably reusing learned constraints; \emph{laziness}, and \emph{compression}.
To evaluate these features, our experiments aim to answer three questions:

\begin{description}
\item[Q2] Can reusing constraints reduce learning times?
\item[Q3] Can laziness reduce learning times?
\item[Q4] Can compression reduce learning times?
\end{description}

\noindent
To answer \textbf{Q2-Q4}, we compare the performance of \name{} with and without these optimisations.

Comparing \name{} against other systems besides \popper{} cannot help us answer questions \textbf{Q1-Q4}.
However, many researchers desire comparisons against `state-of-the-art'.
To appease such a researcher, our experiments try to answer the question:

\begin{description}
\item[Q5] How does \name{} compare against other approaches?
\end{description}
\noindent
To answer \textbf{Q5} we compare \name{} against \popper{}, \metagol{}, \ale{}, and \ilasp{}.
We describe these systems in Section \ref{sec:systems}.
Note that we do not claim that \name{} is \emph{better} than other systems.
All systems have strengths and weaknesses. 
Moreover, as most systems use different biases, a fair comparison is difficult.
Indeed, there will always exist a set of settings whereby system $x$ outperforms system $y$.
Therefore, the reader should not use our experimental results to conclude that system $x$ is better than system $y$.


\subsection{Experimental Domains}

We consider three domains.

\subsubsection{Michalski Trains.}
Michalski trains \cite{michalski:trains} is a classical problem.
The task is to find a hypothesis that distinguishes eastbound trains from westbound trains. 
Figure \ref{fig:trains-prog} shows an example hypothesis that says \emph{a train is eastbound if it has a long carriage with two wheels and another long carriage with three wheels}.
We use this domain because we can easily generate progressively more difficult tasks to test the scalability of the approaches as the solution size grows.
Table \ref{tab:trains-tasks} shows information about the four tasks we consider.
There are 1000 examples but the distribution of positive and negative examples is different for each task.
We randomly sample the examples and split them into 80/20 train/test partitions

\begin{figure}[h]
\begin{center}
\begin{tabular}{l}
\emph{eastbound(A) $\leftarrow$ has\_car(A,B), long(B), two\_wheels(B),}\\
\emph{$\quad\quad\quad\quad\quad\quad\;\;\;\;\;$ has\_car(A,C), three\_wheels(C)}\\
\end{tabular}
\end{center}
\caption{
Target solution for the \emph{trains1} task.
}
\label{fig:trains-prog}
\end{figure}

\begin{table}
\small
\centering
\begin{tabular}{l|c|c|c}
\toprule
\textbf{Task} & \textbf{Num. rules} & \textbf{Num. literals} & \textbf{Max rule size}\\
\midrule
\emph{trains1} & 1 & 6 & 6\\ 
\emph{trains2} & 2 & 11 & 7\\
\emph{trains3} & 3 & 17 & 7\\ 
\emph{trains4} & 4 & 26 & 7\\ 
\bottomrule
\end{tabular}
\caption{
Trains tasks. The values are based on the optimal solution size. The BK contains 27k facts and 20 relations.
}
\label{tab:trains-tasks}
\end{table}

\subsubsection{IGGP.}
In \emph{inductive general game playing} (IGGP) \cite{iggp} agents are given game traces from the general game playing competition \cite{ggp}. 
The task is to induce a set of rules that could have produced these traces.
We use four IGGP games: \emph{minimal decay} (md), \emph{rock, paper, scissors} (rps), \emph{buttons}, and \emph{coins}.
We learn the \emph{next} relation in each game, which is the most difficult to learn \cite{iggp}.


\subsubsection{Program synthesis.}
Inducing complex recursive programs has long been considered a difficult problem \cite{ilp20} and most ILP systems cannot learn recursive programs.
We use the program synthesis dataset introduced by \citet{popper}.


\subsection{Systems}
\label{sec:systems}
To answer \textbf{Q5}, we compare \name{} against \popper{}, \metagol{}, \ale{}, and \ilasp{}.
\begin{description}
\item[\metagol{}] \metagol{} is one of the few systems that can learn recursive Prolog programs.
\metagol{} uses user-provided \emph{metarules} (program templates) to guide the search for a solution.
We use the approximate universal set of metarules described by  \citet{reduce}.
\item[\ale{}] \ale{} excels at learning many large non-recursive rules and should excel at the trains and IGGP tasks.
Although \ale{} can learn recursive programs, it struggles to do so.
\name{} and \ale{} use similar biases so the comparison can be considered reasonably fair.
\item[\ilasp{}] We tried to use \ilasp{}. 
However, \ilasp{} first pre-computes every possible rule in a hypothesis space.
This approach is infeasible for our datasets.
For instance, on the trains tasks, \ilasp{} took 2 seconds to pre-compute rules with three body literals; 20 seconds for rules with four body literals; and 12 minutes for rules with five body literals.  
Since the simplest train task requires rules with six body literals, \ilasp{} is unusable.
In addition, \ilasp{} cannot learn Prolog programs so is unusable in the synthesis tasks.
\end{description}





\subsection{Experimental Results}
We measure predictive accuracy and learning time.
We enforce a timeout of five minutes per task.
We repeat all the experiments\footnote{
The experimental code and data are available at https://github.com/logic-and-learning-lab/aaai22-dcc.
} 20 times and measure the mean and standard deviation.
We use a 3.8 GHz 8-Core Intel Core i7 with 32GB of ram.
All the systems use a single CPU.
    
\begin{table}[ht]
\footnotesize
\centering
\begin{tabular}{l|c|c|c|c}
\toprule
\textbf{Task} & \textbf{\name{}} & \textbf{\popper{}} & \textbf{\ale{}} & \textbf{\metagol{}}\\
\midrule
\emph{trains1} & 100 $\pm$ 0 & 100 $\pm$ 0 & 100 $\pm$ 0 & 27 $\pm$ 0 \\
\emph{trains2} & 98 $\pm$ 0 & 98 $\pm$ 0 & 100 $\pm$ 0 & 19 $\pm$ 0 \\
\emph{trains3} & 98 $\pm$ 0 & 81 $\pm$ 1 & 100 $\pm$ 0 & 79 $\pm$ 0 \\
\emph{trains4} & 100 $\pm$ 0 & 42 $\pm$ 5 & 39 $\pm$ 4 & 32 $\pm$ 0 \\
\midrule
\emph{md} & 99 $\pm$ 0 & 100 $\pm$ 0 & 94 $\pm$ 0 & 11 $\pm$ 0 \\
\emph{buttons} & 98 $\pm$ 0 & 19 $\pm$ 0 & 87 $\pm$ 0 & 19 $\pm$ 0 \\
\emph{rps} & 97 $\pm$ 0 & 18 $\pm$ 0 & 100 $\pm$ 0 & 18 $\pm$ 0 \\
\emph{coins} & 86 $\pm$ 0 & 17 $\pm$ 0 & 17 $\pm$ 0 & 17 $\pm$ 0 \\
\midrule
\emph{dropk} & 99 $\pm$ 0 & 100 $\pm$ 0 & 52 $\pm$ 2 & 50 $\pm$ 0 \\
\emph{droplast} & 100 $\pm$ 0 & 100 $\pm$ 0 & 50 $\pm$ 0 & 50 $\pm$ 0 \\
\emph{evens} & 100 $\pm$ 0 & 100 $\pm$ 0 & 51 $\pm$ 0 & 50 $\pm$ 0 \\
\emph{finddup} & 98 $\pm$ 0 & 98 $\pm$ 0 & 50 $\pm$ 0 & 50 $\pm$ 0 \\
\emph{last} & 100 $\pm$ 0 & 100 $\pm$ 0 & 49 $\pm$ 0 & 55 $\pm$ 3 \\
\emph{len} & 100 $\pm$ 0 & 100 $\pm$ 0 & 50 $\pm$ 0 & 50 $\pm$ 0 \\
\emph{sorted} & 94 $\pm$ 2 & 96 $\pm$ 1 & 70 $\pm$ 1 & 50 $\pm$ 0 \\
\emph{sumlist} & 100 $\pm$ 0 & 100 $\pm$ 0 & 50 $\pm$ 0 & 62 $\pm$ 4 \\
\bottomrule
\end{tabular}
\caption{
Predictive accuracies. 
We round accuracies to integer values. 
The error is standard deviation.
}
\label{tab:q1accs}
\end{table}

\begin{table}[t!]
\small
\centering
\begin{tabular}{l|c|c|c|c}
\toprule
\textbf{Task} & \textbf{\name{}} & \textbf{\popper{}} & \textbf{\ale{}} & \textbf{\metagol{}}\\
\midrule
\emph{trains1} & 8 $\pm$ 2 & 2 $\pm$ 0 & 4 $\pm$ 0.2 & 300 $\pm$ 0 \\
\emph{trains2} & 41 $\pm$ 12 & 7 $\pm$ 0.9 & 1 $\pm$ 0.1 & 300 $\pm$ 0 \\
\emph{trains3} & 106 $\pm$ 17 & 295 $\pm$ 3 & 35 $\pm$ 0.9 & 300 $\pm$ 0 \\
\emph{trains4} & 268 $\pm$ 9 & 295 $\pm$ 2 & 297 $\pm$ 1 & 300 $\pm$ 0 \\
\midrule
\emph{md} & 172 $\pm$ 27 & 52 $\pm$ 1 & 3 $\pm$ 0 & 300 $\pm$ 0 \\
\emph{buttons} & 300 $\pm$ 0 & 299 $\pm$ 0 & 86 $\pm$ 1 & 300 $\pm$ 0 \\
\emph{rps} & 282 $\pm$ 12 & 285 $\pm$ 14 & 4 $\pm$ 0.1 & 0.3 $\pm$ 0 \\
\emph{coins} & 291 $\pm$ 4 & 299 $\pm$ 0 & 300 $\pm$ 0 & 0.4 $\pm$ 0 \\
\midrule
\emph{dropk} & 3 $\pm$ 0.2 & 2 $\pm$ 0.2 & 3 $\pm$ 0.3 & 0.3 $\pm$ 0 \\
\emph{droplast} & 2 $\pm$ 0.2 & 3 $\pm$ 0.1 & 300 $\pm$ 0 & 300 $\pm$ 0 \\
\emph{evens} & 5 $\pm$ 0.4 & 4 $\pm$ 0.1 & 1 $\pm$ 0 & 217 $\pm$ 26 \\
\emph{finddup} & 47 $\pm$ 6 & 13 $\pm$ 0.3 & 1 $\pm$ 0.1 & 300 $\pm$ 0 \\
\emph{last} & 2 $\pm$ 0.4 & 2 $\pm$ 0.1 & 1 $\pm$ 0 & 270 $\pm$ 20 \\
\emph{len} & 16 $\pm$ 2 & 5 $\pm$ 0.1 & 1 $\pm$ 0 & 300 $\pm$ 0 \\
\emph{sorted} & 29 $\pm$ 3 & 19 $\pm$ 1 & 1 $\pm$ 0 & 288 $\pm$ 11 \\
\emph{sumlist} & 18 $\pm$ 0.3 & 19 $\pm$ 0.6 & 0.6 $\pm$ 0 & 225 $\pm$ 29 \\
\bottomrule
\end{tabular}
\caption{
Learning times.
We round times over one second to the nearest second.
The error is standard deviation.
}
\label{tab:q1times}
\end{table}

\begin{table}[h!]
\small
\centering
\begin{tabular}{l|c|c|c|c}
\toprule
 &  & \textbf{Without} & \textbf{Without} &
\textbf{Without}\\
\textbf{task} & \textbf{\name{}} & \textbf{constraints} & \textbf{laziness} &
\textbf{compression}\\
\midrule
\emph{trains1} & 100 $\pm$ 0 & 95 $\pm$ 3 & 100 $\pm$ 0 & 100 $\pm$ 0 \\
\emph{trains2} & 98 $\pm$ 0 & 90 $\pm$ 1 & 96 $\pm$ 1 & 98 $\pm$ 0 \\
\emph{trains3} & 98 $\pm$ 0 & 70 $\pm$ 4 & 97 $\pm$ 0 & 98 $\pm$ 0 \\
\emph{trains4} & 100 $\pm$ 0 & 77 $\pm$ 2 & 97 $\pm$ 0 & 100 $\pm$ 0 \\
\midrule
\emph{md} & 99 $\pm$ 0 & 92 $\pm$ 0 & 95 $\pm$ 1 & 98 $\pm$ 0 \\
\emph{buttons} & 98 $\pm$ 0 & 84 $\pm$ 0 & 96 $\pm$ 0 & 98 $\pm$ 0 \\
\emph{rps} & 97 $\pm$ 0 & 86 $\pm$ 0 & 95 $\pm$ 0 & 97 $\pm$ 0 \\
\emph{coins} & 86 $\pm$ 0 & 77 $\pm$ 4 & 71 $\pm$ 6 & 81 $\pm$ 3 \\
\midrule
\emph{dropk} & 99 $\pm$ 0 & 90 $\pm$ 4 & 99 $\pm$ 0 & 99 $\pm$ 0 \\
\emph{droplast} & 100 $\pm$ 0 & 100 $\pm$ 0 & 100 $\pm$ 0 & 100 $\pm$ 0 \\
\emph{evens} & 100 $\pm$ 0 & 83 $\pm$ 5 & 100 $\pm$ 0 & 100 $\pm$ 0 \\
\emph{finddup} & 98 $\pm$ 0 & 56 $\pm$ 3 & 99 $\pm$ 0 & 97 $\pm$ 2 \\
\emph{last} & 100 $\pm$ 0 & 82 $\pm$ 5 & 100 $\pm$ 0 & 100 $\pm$ 0 \\
\emph{len} & 100 $\pm$ 0 & 85 $\pm$ 5 & 98 $\pm$ 1 & 100 $\pm$ 0 \\
\emph{sorted} & 94 $\pm$ 2 & 60 $\pm$ 2 & 96 $\pm$ 1 & 96 $\pm$ 1 \\
\emph{sumlist} & 100 $\pm$ 0 & 95 $\pm$ 3 & 100 $\pm$ 0 & 100 $\pm$ 0 \\
\bottomrule
\end{tabular}
\caption{
Predictive accuracies.
We round accuracies to integer values.
The error is standard deviation.
}
\label{tab:dccaccs}
\end{table}

\begin{table}[h!]
\centering
\small
\begin{tabular}{l|c|c|c|c}
\toprule
 &  & \textbf{Without} & \textbf{Without} &
\textbf{Without}\\
\textbf{task} & \textbf{\name{}} & \textbf{constraints} & \textbf{laziness} &
\textbf{compression}\\
\midrule
\emph{trains1} & 8 $\pm$ 2 & 38 $\pm$ 20 & 300 $\pm$ 0 & 12 $\pm$ 3 \\
\emph{trains2} & 41 $\pm$ 12 & 285 $\pm$ 14 & 219 $\pm$ 20 & 151 $\pm$ 25 \\
\emph{trains3} & 106 $\pm$ 17 & 300 $\pm$ 0 & 300 $\pm$ 0 & 300 $\pm$ 0 \\
\emph{trains4} & 268 $\pm$ 9 & 300 $\pm$ 0 & 300 $\pm$ 0 & 300 $\pm$ 0 \\
\midrule
\emph{md} & 172 $\pm$ 27 & 300 $\pm$ 0 & 251 $\pm$ 20 & 256 $\pm$ 22 \\
\emph{buttons} & 300 $\pm$ 0 & 300 $\pm$ 0 & 300 $\pm$ 0 & 300 $\pm$ 0 \\
\emph{rps} & 282 $\pm$ 12 & 300 $\pm$ 0 & 300 $\pm$ 0 & 300 $\pm$ 0 \\
\emph{coins} & 291 $\pm$ 4 & 300 $\pm$ 0 & 300 $\pm$ 0 & 300 $\pm$ 0 \\
\midrule
\emph{dropk} & 3 $\pm$ 0.2 & 62 $\pm$ 27 & 20 $\pm$ 2 & 4 $\pm$ 0.5 \\
\emph{droplast} & 2 $\pm$ 0.2 & 4 $\pm$ 0.2 & 101 $\pm$ 2 & 4 $\pm$ 0.2 \\
\emph{evens} & 5 $\pm$ 0.4 & 109 $\pm$ 32 & 75 $\pm$ 3 & 11 $\pm$ 3 \\
\emph{finddup} & 47 $\pm$ 6 & 272 $\pm$ 19 & 178 $\pm$ 15 & 102 $\pm$ 19 \\
\emph{last} & 2 $\pm$ 0.4 & 107 $\pm$ 32 & 46 $\pm$ 3 & 4 $\pm$ 1 \\
\emph{len} & 16 $\pm$ 2 & 95 $\pm$ 30 & 143 $\pm$ 14 & 36 $\pm$ 14 \\
\emph{sorted} & 29 $\pm$ 3 & 286 $\pm$ 14 & 74 $\pm$ 7 & 109 $\pm$ 21 \\
\emph{sumlist} & 18 $\pm$ 0.3 & 50 $\pm$ 19 & 300 $\pm$ 0 & 22 $\pm$ 0.5 \\
\bottomrule
\end{tabular}
\caption{
Learning times.
We round times over one second to the nearest second.
The error is standard deviation.
}
\label{tab:dcctimes}
\end{table}

\subsubsection{Q1. Can \name{} improve predictive accuracies and reduce learning times?}
Table \ref{tab:q1accs} shows the predictive accuracies of \name{} and \popper{}.
The results show that \name{} outperforms \popper{}, especially on the trains and IGGP tasks.
A McNemar's test confirms the significance of the difference between \name{} and \popper{} at the $p < 0.01$ level.
Where \popper{} has low predictive accuracy, it is because it struggled to find a solution in the given time limit and thus returns an empty hypothesis\footnote{
\name{} and \popper{} (and \metagol{}) are all guaranteed to find an optimal solution, if one exists.
They key difference is how long they take to find it.
In this experiment, we use a 5 minute timeout.
However, increasing the timeout does not change the results.
For instance, we repeated one trial of the \emph{buttons} experiment with \textbf{4 hour} timeout.
Even with this large timeout, \popper{} and \metagol{} could still not learn \emph{any} solution.
By contrast, \name{} learns an almost perfect solution in under 5 minutes.
}.

Table \ref{tab:q1times} shows that learning times of \name{} and \popper{}.
A paired t-test confirms that the learning times of \name{} and \popper{} are \emph{not} significantly different.
This result may surprise a reader. 
How can \name{} achieve higher predictive accuracy than \popper{} yet have the same learning time given that it calls \popper{}?
The reason is that \name{} is an anytime algorithm.
On some tasks, \name{} finds the optimal solution early in its search but is unable to prove that it is optimal so continues to search until it reaches the maximum learning time.

\subsubsection{Q2. Can reusing constraints reduce learning times?}
Tables \ref{tab:dccaccs} and \ref{tab:dcctimes} show that reusing constraints is important for high predictive accuracy and low learning times.
A McNemar's test and a paired t-test confirmed the significance of the accuracy and time results respectively at the $p < 0.01$ level.
For instance, without constraint reuse, \name{} takes 107s to learn a \emph{last} solution with 82\% predictive accuracy.
By contrast, with constraint reuse \name{} takes 2s to learn a solution with 100\% accuracy.

\subsubsection{Q3. Can laziness reduce learning times?}
The results show that laziness can drastically reduce learning times.
A paired t-test confirms the significance at the $p < 0.01$ level.
For instance, for the \emph{trains1} task, laziness reduces the learning time from 300s (timeout) to 8s. 
    
\subsubsection{Q4. Can compression reduce learning times?}
The results show that compression can drastically reduce learning times.
A paired t-test confirms the significance at the $p < 0.01$ level.
For instance, for the \emph{sorted} task, compression reduces the learning time from 109s to 29s.

\subsubsection{Q5. How does DCC compare against other ILP systems?
}
The results show that \name{} generally outperforms the other systems in terms of predictive accuracy.
\ale{} performs well on the trains and IGGP tasks but struggles on the program synthesis tasks. 
\popper{} performs well on the trains and program synthesis tasks but struggles on the IGGP tasks.
\metagol{} struggles on most tasks because it uses metarules with at most two body literals.
To learn rules with more than two body literals, \metagol{} must invent new predicates, thus increasing the hypothesis size and hypothesis space.
The most notable difference in accuracies is in the \emph{coins} task.
\name{} achieves 86\% accuracy.
By contrast, the other systems could only achieve the default accuracy of 17\% 





\section{Conclusions and Limitations}
The fundamental challenge in ILP is to efficiently search a large hypothesis space.
To address this challenge, we have introduced an approach called \emph{divide}, \emph{constrain}, and \emph{conquer} (\name{}).
Our approach combines classical divide-and-conquer search with modern constraint-driven search.
Our anytime approach can learn optimal, recursive, and large programs and perform predicate invention.
Our experiments on three domains (classification, inductive general game playing, and program synthesis) show that (i) our approach can drastically improve predictive accuracies and reduce learning times, (ii) reusing learned knowledge is vital for good learning performance, and (iii) our approach can outperform other ILP systems.

\subsection*{Limitations and Future Work}

\paragraph{Noise.}
In contrast to many other systems, such as TILDE and \ale{}, \name{} cannot (explicitly) handle misclassified examples.
However, due to the \dac{} approach, \name{} naturally provides a method to handle misclassified positive examples. 
In future work we want to extend the approach to handle misclassified negative examples.

\paragraph{Expressivity.}
As \name{} uses \popper{} as its underlying search algorithm, it inherits some of the limitations of \popper{}, such as no explicit support for non-observational predicate learning \cite{nonopl}, where a hypothesis must define rules for predicates not seen in the training examples\footnote{Because \popper{} supports predicate invention, the distinction between OPL and non-OPL is unclear as by definition an invented predicate symbol is not given in the examples.},  and a restriction to definite programs.
Future work should address these limitations.

\paragraph{Parallelisation.}
Although multi-core machines are ubiquitous, most ILP approaches are single-core learners, including all the systems mentioned in this paper.
However, our \name{} algorithm is trivially parallelisable as we can independently search for a solution for each chunk.
In future work, we want to explore parallel \name{} approaches.


\subsubsection{Acknowledgments.}
This work was supported by the EPSRC fellowship \emph{The Automatic Computer Scientist} (EP/V040340/1).


\bibliography{arxiv}

\end{document}